%% file: main.tex
\PassOptionsToPackage{section}{placeins}
\documentclass[]{bytedance}
\usepackage[toc,page,header]{appendix}
\usepackage{minitoc}
\usepackage{amsfonts}
\usepackage{amssymb}
\usepackage{tabularx}
\usepackage{listings}
\usepackage{xcolor}
\usepackage{cancel}

\usepackage{tabulary,multirow,xspace}
\usepackage{fixmath,mathtools,nicefrac,mmstyle}
\usepackage{subcaption}
\usepackage{caption}
\usepackage[misc]{ifsym} 
\usepackage{colortbl}

\usepackage{wrapfig}
\usepackage{multicol}
\usepackage[most]{tcolorbox}
\usepackage{pifont}
\usepackage{colortbl, booktabs, multirow, array}

\definecolor{codegreen}{rgb}{0,0.6,0}
\definecolor{codegray}{rgb}{0.5,0.5,0.5}
\definecolor{codepurple}{rgb}{0.58,0,0.82}
\definecolor{backcolour}{rgb}{0.95,0.95,0.92}
\definecolor{boxblue}{RGB}{57,89,163}
\definecolor{boxbluebg}{RGB}{230,237,250} 

\lstdefinestyle{mystyle}{
    backgroundcolor=\color{backcolour},   
    commentstyle=\color{codegreen},
    keywordstyle=\color{magenta},
    numberstyle=\tiny\color{codegray},
    stringstyle=\color{codepurple},
    basicstyle=\ttfamily\footnotesize,
    breakatwhitespace=false,         
    breaklines=true,                 
    captionpos=b,                    
    keepspaces=true,                 
    numbers=none,                    
    numbersep=5pt,                  
    showspaces=false,                
    showstringspaces=false,
    showtabs=false,                  
    tabsize=2
}
\definecolor{mygray1}{gray}{.95}
\definecolor{mygray2}{gray}{.9}
\definecolor{mygray3}{gray}{.95}
\usepackage{pifont}

\newlength\savewidth
\newcolumntype{x}[1]{>{\centering\arraybackslash}p{#1pt}}

\newcommand{\app}{\raise.17ex\hbox{$\scriptstyle\sim$}}

\usepackage{xcolor}
\input{macros}

\usepackage{listings}

\usepackage{algorithm}
\usepackage{algorithmic}
\definecolor{myblue}{RGB}{210, 225, 255}
\definecolor{mytextblue}{RGB}{51, 161, 201}
\definecolor{mypurple}{RGB}{218, 112, 214}

\definecolor{commentgreen}{rgb}{0.1, 0.4, 0.1}
\definecolor{keywordblue}{rgb}{0.1, 0.1, 0.7}
\definecolor{stringred}{rgb}{0.7, 0.1, 0.1}

\lstdefinestyle{mystyle}{
    commentstyle=\color{commentgreen},
    keywordstyle=\color{keywordblue},   
    stringstyle=\color{stringred},
    basicstyle=\ttfamily\scriptsize, 
    breaklines=true,
    keepspaces=true,
    showstringspaces=false,
    frame=none,                     
    language=Python, 
}

\newcommand{\name}{Stream-R1}
\title{\name{}: Reliability-Perplexity Aware Reward Distillation for Streaming Video Generation}

\author{
Bin Wu$^1$\quad
Mengqi Huang$^{1,\dagger,\ddagger}$\quad
Shaojin Wu$^{3,\ddagger}$\quad
\\
Weinan Jia$^1$\quad
Yuxin Wang$^2$\quad
Zhendong Mao$^1$\quad
Yongdong Zhang$^1$
}
\affiliation{
$^1$University of Science and Technology of China\quad
$^2$FrameX.AI\quad
$^3$Independent Researcher \\[2pt]
{\normalfont\small $^\dagger$Corresponding author\quad $^\ddagger$Project lead}
}

\input{sec/1_abs}

\date{\today}
\checkdata[Project Page]{\url{https://stream-r1.github.io}}
\correspondence{Mengqi Huang at \email{huangmq@ustc.edu.cn}}

\begin{document}
\maketitle

\input{sec/2_intro}
\input{sec/3_related}
\input{sec/4_methods}

\input{sec/5_exps}
\input{sec/7_conclusion}

\clearpage

\bibliographystyle{plainnat}
\bibliography{main}

\end{document}

%% file: macros.tex
\usepackage{graphicx}
\usepackage{amssymb}
\usepackage{pifont}
\usepackage{floatrow}
\usepackage{amsmath} 
\usepackage{float}
\usepackage{wrapfig}
\usepackage{multirow}
\usepackage{tcolorbox}
\tcbuselibrary{breakable, skins, raster}
\usepackage{listings}

%% file: sec/1_abs.tex
\abstract{
Distillation-based acceleration has become the foundational technique 
for making autoregressive streaming video diffusion models practical, 
with distribution matching distillation as the de facto choice. 
However, existing methods train the student to match the teacher's 
output in an indiscriminative manner, treating every rollout, every 
frame, and every pixel as equally reliable supervision. We argue that 
this indiscriminative treatment caps the upper bound of distilled 
quality because it overlooks two complementary axes of variance in 
the DMD supervision signal: \textit{Inter-Reliability} across 
different student rollouts on which the supervision varies in 
reliability, and \textit{Intra-Perplexity} across spatial regions 
and temporal frames that contribute unequally to where the current 
quality can still be improved. The distillation objective thus 
implicitly conflates two distinct questions under a single uniform 
weight: whether to learn from each rollout, and where to concentrate 
optimization within each rollout. To address this, we propose 
Stream-R1, a Reliability-Perplexity Aware Reward Distillation 
framework that adaptively reweights the distillation objective at 
both the rollout level and the spatiotemporal-element level through 
a single shared reward-guided mechanism. At the Inter-Reliability 
level, Stream-R1 rescales each rollout's loss by an exponential of 
a pretrained video reward score, so that rollouts on which the DMD 
supervision is reliable dominate the gradient signal. At the 
Intra-Perplexity level, it back-propagates the same reward model 
to extract per-pixel gradient saliency, which is factored into 
spatial and temporal weights that concentrate optimization pressure 
on the regions and frames where further refinement yields the 
largest expected gain. An adaptive balancing mechanism further 
prevents any single quality axis from dominating across visual 
quality, motion quality, and text alignment. Stream-R1 attains 
consistent improvements on all three quality dimensions over 
distillation baselines on standard streaming video generation 
benchmarks, without architectural modification to the student and 
at no additional inference cost.
}

%% file: sec/2_intro.tex
\section{Introduction}
\label{sec:intro}
\begin{figure}
    \centering
    \includegraphics[width=1\linewidth]{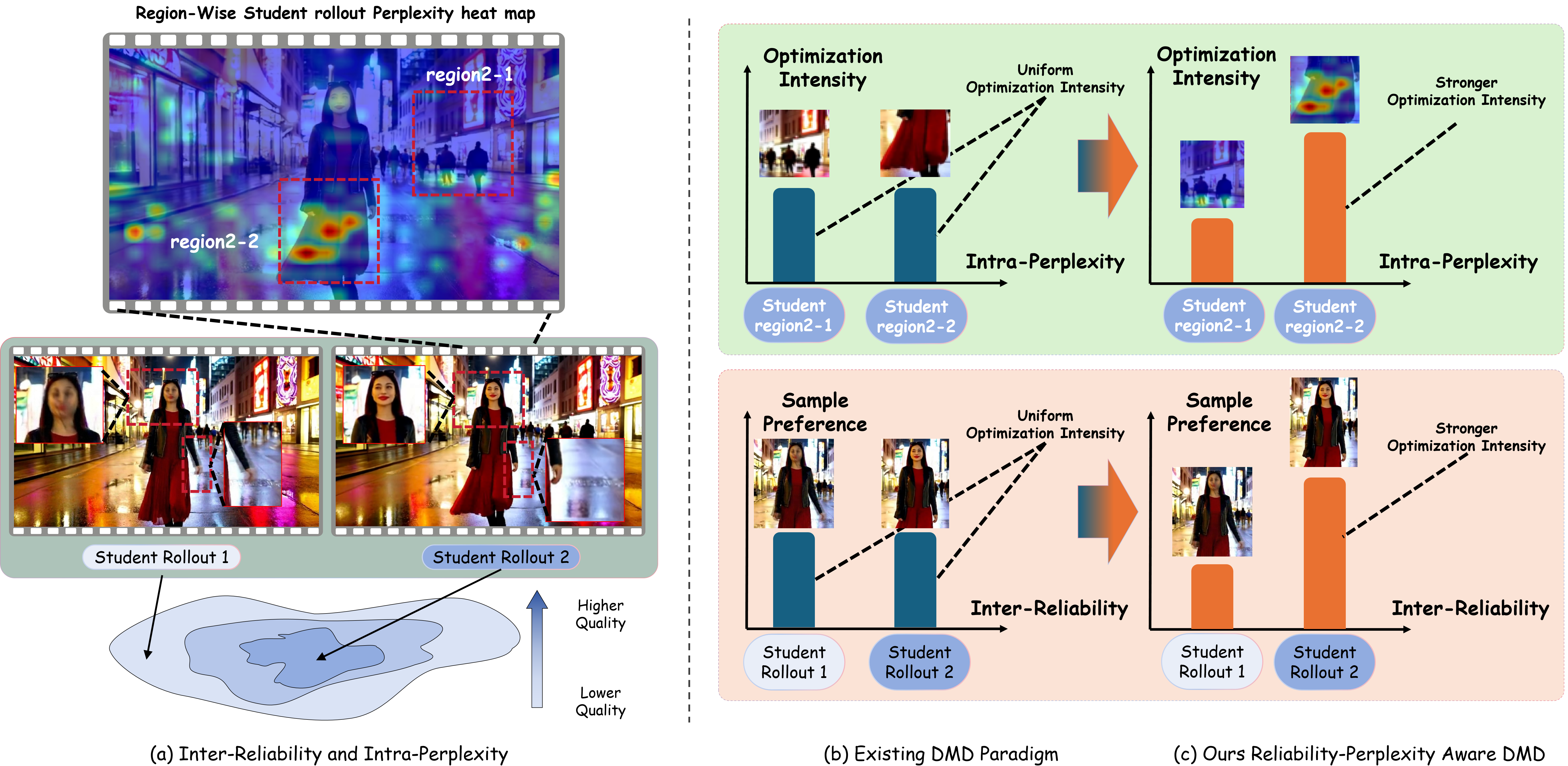}
    \caption{\textbf{Motivation of Stream-R1.} (a) The DMD supervision
    signal exhibits two complementary axes of variance:
    \textit{Inter-Reliability} across different rollouts, and
    \textit{Intra-Perplexity} across spatiotemporal regions within
    each rollout. (b) The existing DMD paradigm assigns uniform sample
    preference to all rollouts and uniform optimization intensity to
    all regions, regardless of their reliability or perplexity. (c)
    Our reliability-perplexity aware DMD upweights rollouts on which
    the supervision is reliable and concentrates higher optimization
    intensity on regions where further refinement yields the largest
    expected gain, all driven by a single reward model.}
    \label{fig:motivation}
\end{figure}

Recent advances in video diffusion
models~\cite{zheng2024open,polyak2024movie,yang2024cogvideox,kong2024hunyuanvideo}
have driven text-to-video generation to unprecedented visual quality.
However, their reliance on multi-step denoising over a fixed temporal
window imposes prohibitive inference cost and precludes streaming
interactivity as well as scalable long-video synthesis. Autoregressive
streaming video diffusion
models~\cite{yin2025slow,huang2025self,chen2025skyreels,chen2024diffusion}
have emerged as a promising remedy, converting bidirectional
architectures into causal generators that produce frames sequentially
and, in principle, support unbounded video generation. To further make
this paradigm practical, distillation-based
acceleration~\cite{yin2024one,lin2025autoregressive,yang2025longlive}
compresses the expensive multi-step teacher into an efficient few-step
student, with distribution matching distillation
(DMD)~\cite{yin2024one} emerging as the de facto choice. Despite their
diverse designs, these distillation-driven streaming video generation
methods all revolve around the same key challenge: \textit{how to
effectively align the student's output distribution with the
high-quality mode of a multi-step teacher's distribution}, so that the
student can inherit the teacher's generative fidelity while operating
under a causal, streaming regime.

Existing efforts toward this goal can be broadly organized into two
complementary directions. The first augments the distillation
\textit{objective} with additional supervisory signals:
DMD2~\cite{yin2024improved} introduces a GAN discriminator trained on
real videos to compensate for the mode-covering bias of the teacher's
score. The second reshapes the \textit{rollout} on which distillation
is performed: Self-Forcing~\cite{huang2025self} trains the student on
its own autoregressive rollouts to close the train-test distribution
gap, while LongLive~\cite{yang2025longlive} further scales this idea
to minute-long generation through memory mechanisms and chunk-level
objectives. Despite differing in \textit{where} they intervene, these
approaches share a fundamental commonality: they all minimize the
per-instance distribution discrepancy between student and teacher
outputs in an \textit{indiscriminative} manner. Every rollout, every
frame, and every pixel is matched against the teacher with equal
weight, and the distillation objective implicitly treats the
supervision signal on every element as equally reliable.

We argue that this paradigm of indiscriminative distillation
inherently overlooks two complementary axes of variance in the DMD
supervision signal, as illustrated in
Fig.~\ref{fig:motivation}(a): \textit{Inter-Reliability}, the
variation in supervision reliability across different student
rollouts, and \textit{Intra-Perplexity}, borrowing the term from language
modeling to denote the variation across spatiotemporal regions in
how much further refinement can still improve the underlying
quality within each individual rollout.
\textit{Inter-Reliability} arises because the DMD gradient
$g = f_{\text{fake}} - f_{\text{real}}$ is itself an estimate, and
its reliability varies substantially across student rollouts. The
teacher-derived $f_{\text{real}}$ is fundamentally a conditional
denoiser rather than a generator: it provides a local correction
whose direction is determined by where the input already lies, not by
where high-quality samples globally reside. When a student rollout
already lies near the teacher's high-quality mode, $f_{\text{real}}$
produces a correction that points within that mode and $g$ faithfully
reflects the residual gap that the student should close. When a
rollout falls far from this mode, $f_{\text{real}}$ can only produce
a correction toward the low-quality region the sample originated
from, and $g$ on such rollouts encodes a within-low-quality
refinement rather than a path toward the high-quality mode. The
online-trained $f_{\text{fake}}$ exhibits an analogous dependence on
the student's current distribution. Existing DMD methods average $g$
with equal weight across all rollouts, conflating these two regimes
and diluting the fraction of supervision that genuinely points
toward the high-quality mode.
\textit{Intra-Perplexity}, in contrast, arises because within a
single rollout different spatial regions and temporal frames
contribute unequally to where the current quality can still be
improved. Some regions still lie far from the high-quality mode and yield
large quality gains under further refinement, while others have
already approached this mode locally and yield diminishing returns.
As shown in Fig.~\ref{fig:motivation}(b), existing methods apply an
indiscriminative loss across all pixels and frames, spending
optimization budget on regions where the reward has already saturated
while leaving high-perplexity regions under-supervised. Taken
together, these two axes suggest that the distillation objective
should not be governed by a single uniform weight, but rather by two
complementary questions: \textit{whether} the supervision on each
rollout is reliable enough to learn from, and \textit{where} to
concentrate optimization within each rollout.

Guided by these two questions, we propose Stream-R1, illustrated in
Fig.~\ref{fig:motivation}(c), a reliability-perplexity aware
distribution matching distillation framework that adaptively
reweights the DMD objective at both the rollout level and the
spatiotemporal-element level through a single reward-guided
mechanism. At the \textit{Inter-Reliability} level, Stream-R1
evaluates each rollout with a pretrained video reward model and
rescales its distillation loss by an exponential of the resulting
score, so that rollouts on which the DMD supervision is reliable
dominate the gradient signal. At the \textit{Intra-Perplexity}
level, Stream-R1 back-propagates the same reward model to obtain a
per-pixel gradient saliency volume, which serves as a perplexity
signal: regions with higher saliency correspond to content where the
reward score is currently most sensitive to small perturbations,
indicating that the local reward landscape has not yet flattened. The
saliency is factorized into spatial and temporal components and
composed into a per-element weighting on the DMD loss, concentrating
optimization pressure where further refinement yields the largest
expected gain. To prevent any single quality dimension from
dominating the supervision, both the Inter-Reliability score and the
Intra-Perplexity saliency aggregate three complementary axes—visual
quality, motion quality, and text alignment—and are adaptively fused
according to the current improvement trajectory of each axis. As a
result, Stream-R1 retains the tractability of the DMD objective
while replacing its uniform weighting with reliability-perplexity
aware guidance that requires no architectural change to the student
and adds no cost at inference time.

\textbf{Conceptual contribution.} We reformulate DMD-based
distillation for autoregressive streaming video generation as a
\textit{reliability-perplexity aware} process. We identify that
prevailing methods match every rollout, every frame, and every pixel
against the teacher with equal weight, and we argue that this
indiscriminative treatment overlooks two complementary axes of
variance in the DMD supervision signal: Inter-Reliability across
rollouts and Intra-Perplexity within each rollout. Both axes must be
addressed for the student to converge toward the teacher's
high-quality mode.

\textbf{Technical contribution.} We instantiate this formulation as
Stream-R1, a unified reward-guided framework that derives both an
Inter-Reliability weight and an Intra-Perplexity weight from a single
pretrained video reward model, with adaptive balancing across visual
quality, motion quality, and text alignment. Stream-R1 attains
consistent improvements on all three quality dimensions over
DMD-based baselines on standard streaming video generation
benchmarks, without any architectural modification to the student
and at no additional inference cost.

%% file: sec/3_related.tex
\section{Related Work}
\label{sec:Related Work}

\subsection{Streaming Video Generation}
Video diffusion models~\cite{wan2025wan,yang2024cogvideox,kong2024hunyuanvideo,
hacohen2024ltx} have achieved remarkable results in visual synthesis, yet their 
reliance on multi-step denoising over fixed-length temporal windows limits both 
inference efficiency and temporal scalability. To overcome these constraints, a 
growing body of work reformulates video generation as autoregressive diffusion, 
enabling streaming, frame-by-frame synthesis that can in principle extend to 
arbitrary temporal horizons~\cite{chen2024diffusion,gao2025longvie,
henschel2025streamingt2v,li2025stable,zhang2025frame,cui2025self}. 
Pyramidal-Flow~\cite{jin2024pyramidal} employs multi-scale flow matching to 
reduce the computational burden of long sequences; SkyReels-V2~\cite{chen2025skyreels} 
integrates diffusion forcing with structural planning for scalable synthesis; 
FAR~\cite{gu2025long} combines short- and long-term contexts via flexible 
positional encoding; and MAGI-1~\cite{teng2025magi} adopts chunk-wise prediction 
for scalable autoregressive generation. A complementary line of work accelerates inference through distillation. 
Distribution matching distillation (DMD)~\cite{yin2024one} compresses multi-step 
teacher inference into few-step student generation by minimizing their output 
distribution divergence. CausVid~\cite{lin2025autoregressive} extends this 
framework to causal video generation by reformulating bidirectional diffusion as 
autoregressive generation through distribution matching. Self-Forcing~\cite{huang2025self} 
further addresses the train--test discrepancy in autoregressive distillation by 
feeding the model's own predictions as context during training rather than 
ground-truth latents. LongLive~\cite{yang2025longlive} extends this paradigm 
through KV recaching and stream-based fine-tuning for long video generation, 
while Rolling-Forcing~\cite{liu2025rolling} introduces joint denoising for 
simultaneous multi-frame processing. Despite significant advances in efficiency and temporal extent,
these methods all learn from the teacher in an indiscriminative
manner, applying uniform optimization pressure to every rollout,
every spatial region, and every temporal frame. This treatment
overlooks two sources of variance in the DMD supervision signal:
across rollouts, the gradient varies in how reliably it points
toward the teacher's high-quality mode; within each rollout,
spatial regions and temporal frames vary in how much further
refinement can still raise the quality.

\subsection{Reinforcement Learning for Visual Generation}

Reinforcement learning (RL) has emerged as a principled framework for optimizing non-differentiable objectives and aligning generative models with human preferences, achieving transformative success in large language models~\cite{ouyang2022training,schulman2017proximal,rafailov2023direct,guo2025deepseek} and increasingly in visual generation~\cite{black2023training,xue2025dancegrpo}. Several efforts focus on building specialized reward models and preference datasets for visual content. VideoReward~\cite{liu2025improving}, VideoScore~\cite{he2024videoscore}, and VisionReward~\cite{xu2026visionreward} provide multi-dimensional quality scores spanning visual fidelity, motion coherence, and semantic alignment, serving as optimization targets for downstream training. On the algorithmic side, direct preference optimization (DPO) has been extended from language models to image~\cite{wallace2024diffusion,jiang2025distribution} and video~\cite{liu2025videodpo} diffusion models, learning directly from pairwise preference data without explicit reward modeling. Policy gradient methods such as Flow-GRPO~\cite{liu2025flow} adapt group relative policy optimization to flow matching, enabling online RL fine-tuning for improved compositional accuracy. Reward Forcing~\cite{lu2025reward} combines reward feedback with distribution matching distillation, reweighting the distillation loss by the exponential of a scalar reward to bias the student toward higher-quality regions of the generation manifold. Whereas prior reward-guided methods primarily use the reward to
fine-tune the generator end-to-end or to filter training data, our
work brings the reward signal directly into the DMD distillation
objective at two complementary levels: an Inter-Reliability scalar
weight that modulates each rollout's contribution to the loss, and
an Intra-Perplexity per-element weight derived from the reward
gradient that concentrates optimization on regions and frames where
further refinement yields the largest expected gain.

%% file: sec/4_methods.tex
\section{Methodology}

\begin{figure}[ht]
    \centering
    \includegraphics[width=1\linewidth]{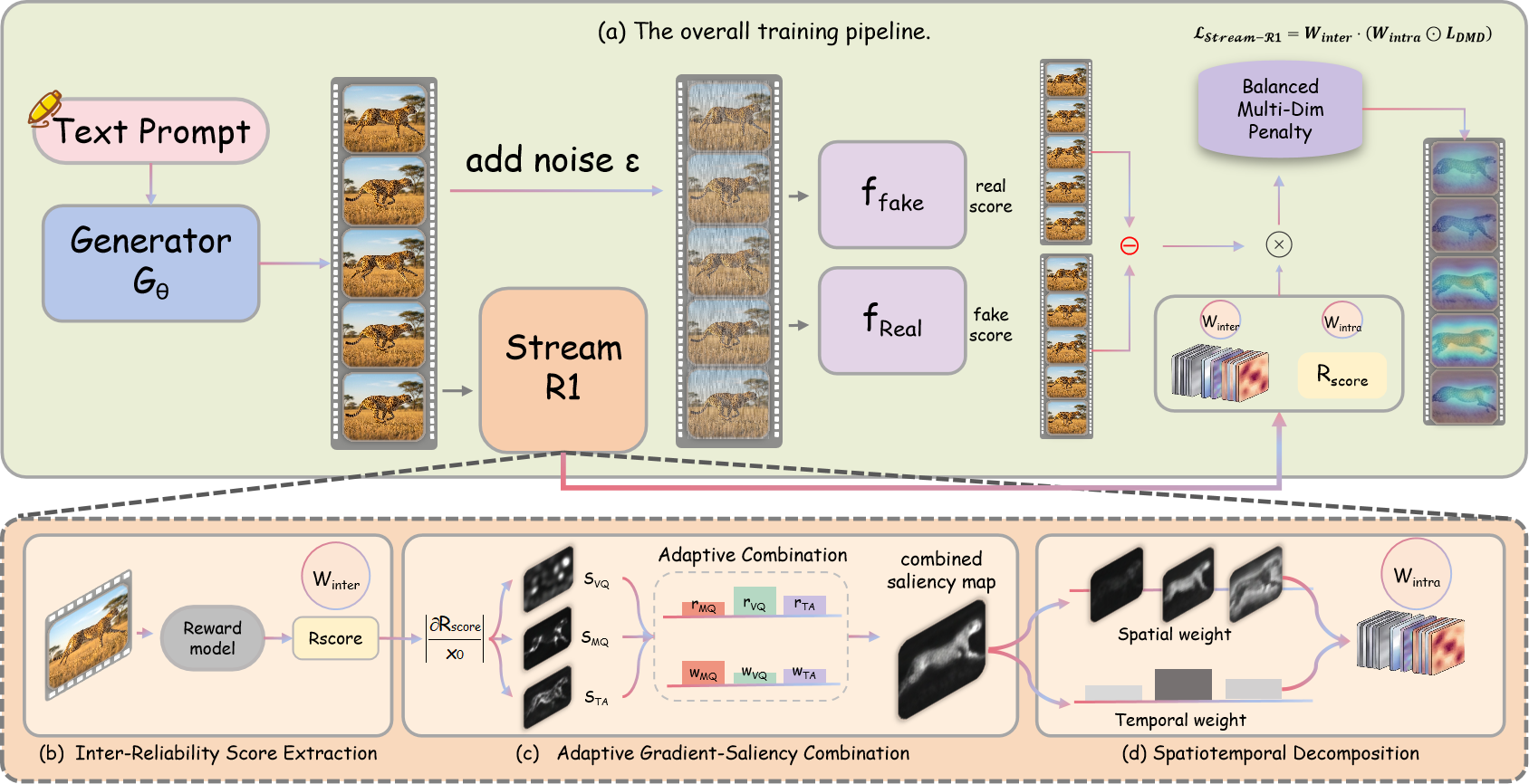}
    \caption{\textbf{Overview of Stream-R1.}
\textbf{(a)} The fake rollout from $G_\theta$ is scored by DMD
networks $f_{\text{fake}}, f_{\text{Real}}$ and the Stream R1 module;
the distillation signal is modulated by an Inter-Reliability weight
$w_{\text{inter}}$ and an intra-instance weight $\mathbf{W}_{\text{intra}}$ to
form $\mathcal{L}_{\text{Stream-R1}} = \mathbf{W}_{\text{inter}} \cdot
(W_{\text{intra}} \odot \mathcal{L}_{\text{DMD}})$.
\textbf{Bottom:} Inside the Stream R1 module,
\textbf{(b)~Inter-Reliability Score Extraction} produces a scalar reward
$R_{\text{score}}$ for $w_{\text{inter}}$ and per-axis saliencies
$s_{\text{VQ/MQ/TA}}$;
\textbf{(c)~Adaptive Gradient-Saliency Combination} fuses the three saliencies
into a unified map;
\textbf{(d)~Spatiotemporal Decomposition} factorizes the map into
spatial and temporal weights to form $W_{\text{intra}}$.
A single reward model drives both weights.}

    \label{fig:framework}
\end{figure}
\label{sec:Methodology}
We first introduce the preliminaries on reward-guided video distillation. We then present the four key components of Stream-R1 in turn: Inter-Reliability score extraction in Sec.~\ref{sec:sub3.2}, adaptive gradient-saliency combination in Sec.~\ref{sec:sub3.3}, spatiotemporal saliency decomposition in Sec.~\ref{sec:sub3.4}, and balanced multi-dimensional reward in Sec.~\ref{sec:sub3.5}. An overview of Stream-R1 is illustrated in Fig.~\ref{fig:framework}.

\subsection{Preliminary}
\label{sec:sub3.1}

\textbf{Video Diffusion Distillation.}
Given a pretrained video diffusion teacher $\boldsymbol{\epsilon}_\theta$, distillation methods train a student generator $G_\phi$ to produce high-quality videos in significantly fewer denoising steps. In the distribution matching distillation (DMD) framework, the student learns to match the output distribution of the teacher by minimizing a KL-divergence-based objective. Concretely, given a text prompt $c$, the student generates a clean latent $\mathbf{x}_0 = G_\phi(c)$. A noisy version $\mathbf{x}_t$ is constructed by adding noise at a randomly sampled timestep $t$, and a pair of critic networks $f_{\text{real}}$ and $f_{\text{fake}}$ estimate the score functions of the real and fake distributions, respectively. The distillation gradient is computed as:
\begin{equation}
    \mathbf{g} = f_{\text{fake}}(\mathbf{x}_t, c) - f_{\text{real}}(\mathbf{x}_t, c),
    \label{eq:kl_grad}
\end{equation}
and the base distillation loss takes the form:
\begin{equation}
    \mathcal{L}_{\text{DMD}} = \frac{1}{2}\left\| \mathbf{x}_0 - \text{sg}\!\left(\mathbf{x}_0 - \hat{\mathbf{g}}\right) \right\|^2,
    \label{eq:dmd_loss}
\end{equation}
where $\hat{\mathbf{g}}$ denotes the normalized gradient and $\text{sg}(\cdot)$ is the stop-gradient operator.

\subsection{Inter-Reliability Weighting}
\label{sec:sub3.2}

In DMD, the student is supervised by the gradient
$g = f_{\text{fake}} - f_{\text{real}}$ on each generated rollout,
but $g$ is itself an estimate whose reliability varies substantially
across rollouts. The teacher-derived $f_{\text{real}}$ is
fundamentally a conditional denoiser: it provides a local correction
whose direction is determined by where the input already lies, not
by where high-quality samples globally reside. When a student
rollout already lies near the teacher's high-quality mode,
$f_{\text{real}}$ produces a correction that points within that
mode and $g$ faithfully reflects the residual gap the student
should close. When a rollout falls far from this mode,
$f_{\text{real}}$ can only produce a correction toward the
low-quality region the sample originated from, and $g$ on such
rollouts encodes a within-low-quality refinement rather than a
path toward the high-quality mode. The online-trained
$f_{\text{fake}}$ exhibits an analogous dependence on the student's
current distribution. Existing DMD methods average $g$ with equal
weight across all rollouts, conflating these two regimes and
diluting the fraction of supervision that genuinely points toward
the high-quality mode. We address this Inter-Reliability variance
by assigning each rollout a per-sample loss multiplier that grows
with its overall reward, so that rollouts on which the DMD
supervision is reliable contribute more strongly while those
encoding only within-low-quality refinement are attenuated.

Concretely, we query a pretrained video reward model on the
student-generated rollout $\mathbf{V}$ and aggregate its
per-dimension scalar rewards $\{r_d\}_{d \in \mathcal{D}}$ into a
single balanced overall reward $r_{\text{final}}$, as defined in
Eq.~\eqref{eq:balanced_reward}. The reward score serves as a proxy
for supervision reliability: rollouts in the reward model's
high-scoring region lie within the teacher's high-quality mode
where $f_{\text{real}}$ has been densely trained and the student
distribution has stabilized, so $g$ on these rollouts more
faithfully reflects the true KL gradient. We convert this scalar
into a per-sample loss multiplier through an exponential
reweighting:
\begin{equation}
    \mathbf{W}_{\text{inter}} = \exp(\beta \cdot r_{\text{final}}),
    \label{eq:winter}
\end{equation}
where $\beta > 0$ is a temperature controlling the sharpness of
the reweighting. Because the exponential is monotonically
increasing in $r_{\text{final}}$, rollouts on which $g$ is reliable
dominate the gradient signal, biasing the optimizer toward updates
supported by accurate score estimates rather than within-low-quality
refinements.

\subsection{Adaptive Gradient-Saliency Combination}
\label{sec:sub3.3}

The Inter-Reliability weight $\mathbf{W}_{\text{inter}}$ accounts
for variance \textit{across} rollouts, but it leaves variance
\textit{within} each individual rollout unaddressed. Different
spatial regions and temporal frames within the same rollout
contribute unequally to where the current quality can still be
improved: some regions are far from the high-quality mode and yield large gains under further refinement, while
others have already approached their local optimum and yield
diminishing returns. Applying a uniform per-element loss across
all pixels and frames therefore wastes optimization budget on
regions where the reward has already saturated and under-supervises
regions with substantial improvement potential. We address this
Intra-Perplexity variance by deriving a per-element weight that
localizes optimization pressure on the spatiotemporal regions where
further refinement yields the largest expected gain.

A natural source of such localization is the reward model itself.
When the model evaluates a generated video, each input pixel
contributes differently to the local reward landscape, and the
gradient of the score with respect to the input naturally encodes
this contribution. Regions with large gradient magnitudes are
those where the reward score is currently most sensitive to small
perturbations, indicating both that the reward landscape has not
yet flattened in that region and that targeted optimization there
would most significantly raise the quality. Existing reward-guided
distillation treats the reward as an opaque scalar and discards
this rich spatial and temporal information; we recover it by
back-propagating through the reward model.

Formally, given the student-generated video
$\mathbf{V} \in \mathbb{R}^{F \times H \times W \times 3}$ and a
quality dimension $d \in \{\text{VQ}, \text{MQ}, \text{TA}\}$, the
reward model $R_d$ maps $\mathbf{V}$ to a scalar score
$r_d = R_d(\mathbf{V})$. We compute the per-axis saliency map by
back-propagating through $R_d$ and taking the absolute gradient
with respect to the input pixels:
\begin{equation}
    \mathbf{S}^{(d)} = \left| \frac{\partial R_d(\mathbf{V})}{\partial \mathbf{V}} \right| \in \mathbb{R}^{F \times H \times W},
    \label{eq:saliency}
\end{equation}
where the absolute value aggregates positive and negative
sensitivities into a unified magnitude of local reward
sensitivity. This computation requires only a single backward pass
through the reward model per quality dimension, introducing
negligible overhead relative to the diffusion model's own forward
and backward passes.

Different quality dimensions assess complementary aspects of the
generated video, and their saliency maps naturally highlight
different spatiotemporal regions. Combining these per-dimension
maps into a unified guide is therefore essential for comprehensive
quality optimization. We adopt an adaptive combination strategy
that dynamically adjusts the contribution of each dimension based
on its current reward score, allocating proportionally greater
attention to dimensions with lower scores that exhibit larger room
for improvement:
\begin{equation}
    \alpha_d = \frac{\exp(-r_d / \tau)}{\sum_{d'} \exp(-r_{d'} / \tau)}, \qquad
    \mathbf{S}_{\text{combined}} = \sum_{d} \alpha_d \cdot \mathbf{S}^{(d)},
    \label{eq:combine}
\end{equation}
where $r_d$ is the current scalar reward for dimension $d$ and
$\tau$ is a temperature parameter controlling the sharpness of the
allocation. When $\tau \to 0$, the combination reduces to selecting
the saliency map of the worst-performing dimension; when
$\tau \to \infty$, it degenerates to uniform averaging. In
practice, moderate values of $\tau$ yield a smooth blend that
prioritizes the weakest dimension while retaining cues from all
dimensions, enabling the optimizer to address multiple quality
deficiencies through a single unified saliency map.

\subsection{Spatiotemporal Saliency Decomposition}
\label{sec:sub3.4}

The combined saliency volume
$\mathbf{S}_{\text{combined}} \in \mathbb{R}^{F \times H \times W}$
jointly encodes both within-frame spatial structure and
across-frame temporal structure of reward sensitivity. Directly
normalizing this volume globally would entangle these two factors:
a frame with globally high saliency would dominate the weight map
regardless of its internal spatial structure. To disentangle these
effects, we propose a factored decomposition that separately
normalizes the spatial and temporal components before composing
them into a unified weight $\mathbf{W}_{\text{intra}}$.

\textbf{Temporal Weight Extraction.}
We extract the per-frame saliency by averaging over spatial
dimensions:
\begin{equation}
    \mathbf{p}_f = \frac{1}{HW}\sum_{h,w} \mathbf{S}_{\text{combined}}[f, h, w], \qquad f \in \{1, \dots, F\}.
    \label{eq:temporal_profile}
\end{equation}
The resulting temporal profile $\mathbf{p} \in \mathbb{R}^{F}$ is
then normalized via min-max scaling and clamped to a minimum
weight $\tau_{\min}$ to prevent any frame from being entirely
suppressed:
\begin{equation}
    \hat{p}_f = \frac{p_f - p_{\min}}{p_{\max} - p_{\min}}, \qquad
    w_f^{(t)} = \max\!\big(\hat{p}_f,\; \tau_{\min}\big).
    \label{eq:temporal_weight}
\end{equation}
The temporal weights are then mean-normalized such that
$\frac{1}{F}\sum_f w_f^{(t)} = 1$, ensuring that the overall loss
magnitude is preserved.

\textbf{Spatial Weight Extraction.}
For spatial weights, we perform per-frame normalization
independently, so that each frame's internal spatial structure is
preserved regardless of its global saliency magnitude:
\begin{equation}
    \hat{s}_{f,h,w} = \frac{s_{f,h,w} - s_f^{\min}}{s_f^{\max} - s_f^{\min}}, \qquad
    w_{f,h,w}^{(s)} = \max\!\big(\hat{s}_{f,h,w},\; \sigma_{\min}\big),
    \label{eq:spatial_weight}
\end{equation}
where $s_f^{\min}$ and $s_f^{\max}$ denote the minimum and maximum
saliency values within frame $f$. Each frame's spatial weights are
independently mean-normalized to $1$.

\textbf{Composition.}
The final per-element weight map is obtained by multiplying the
temporal and spatial components, followed by a global
mean-normalization:
\begin{equation}
    \mathbf{W}_{\text{intra}}[f, h, w] = \frac{w_f^{(t)} \cdot w_{f,h,w}^{(s)}}{\frac{1}{FHW}\sum_{f',h',w'} w_{f'}^{(t)} \cdot w_{f',h',w'}^{(s)}}.
    \label{eq:final_weight}
\end{equation}
This factored design offers two advantages. First, it allows the
temporal and spatial components to operate at different
granularities: the temporal weights modulate the contribution of
entire frames, while the spatial weights refine the within-frame
distribution. Second, the independent per-frame spatial
normalization ensures that every frame retains a meaningful
internal contrast, even if its overall saliency magnitude is low.

\subsection{Balanced Multi-Dimensional Reward}
\label{sec:sub3.5}

When optimizing across multiple quality dimensions simultaneously,
a na\"ive summation of rewards risks unbalanced improvement, where
the optimizer disproportionately focuses on whichever dimension
yields the easiest gains while neglecting others. To address this,
we introduce a balance penalty that discourages divergent
improvement trajectories across dimensions.

We maintain a sliding window of size $N$ tracking the per-dimension
reward history. At each optimization step, we compute the
improvement for each dimension $d$ as the difference between its
recent and baseline average rewards:
\begin{equation}
    \Delta_d = \bar{r}_d^{\text{recent}} - \bar{r}_d^{\text{baseline}},
    \label{eq:delta}
\end{equation}
where $\bar{r}_d^{\text{baseline}}$ and $\bar{r}_d^{\text{recent}}$
are computed from the first and second halves of the history
window, respectively. The balance penalty is defined as the
standard deviation of improvements across dimensions:
\begin{equation}
    \mathcal{P}_{\text{bal}} = \text{std}\!\left(\{\Delta_d\}_{d \in \mathcal{D}}\right),
    \label{eq:balance_penalty}
\end{equation}
which is subtracted from the base reward with a weighting
coefficient $\lambda$:
\begin{equation}
    r_{\text{final}} = \frac{1}{|\mathcal{D}|}\sum_{d} r_d - \lambda \cdot \mathcal{P}_{\text{bal}}.
    \label{eq:balanced_reward}
\end{equation}
The penalty is activated only after a warmup period of $N$ steps
to allow initial reward estimates to stabilize. Two mechanisms
jointly enforce balanced improvement: the per-dimension softmax
weights $\alpha_d$ in Eq.~\eqref{eq:combine} directly redistribute
saliency toward dimensions with currently lower scores, while
the standard-deviation penalty $\mathcal{P}_{\text{bal}}$ globally
suppresses the effective reward whenever improvement trajectories
diverge across dimensions. Together they discourage the optimizer
from over-fitting any single quality axis.

\subsection{Overall Objective}
\label{sec:sub3.6}

Combining all components, the final Stream-R1 generator loss is:
\begin{equation}
    \mathcal{L}_{\text{Stream-R1}} = \frac{1}{2}\, \mathbf{W}_{\text{inter}} \cdot \text{mean}\!\left(\mathbf{W}_{\text{intra}} \odot \left\| \mathbf{x}_0 - \text{sg}(\mathbf{x}_0 - \hat{\mathbf{g}}) \right\|^2\right),
    \label{eq:full_loss}
\end{equation}
where $\mathbf{W}_{\text{inter}}$ is the Inter-Reliability weight
from Eq.~\eqref{eq:winter},
$\mathbf{W}_{\text{intra}} \in \mathbb{R}^{F \times H \times W}$
is the Intra-Perplexity weight map from
Eq.~\eqref{eq:final_weight}, and $\odot$ denotes element-wise
multiplication. The weight map $\mathbf{W}_{\text{intra}}$ is
broadcast across the channel dimension to match the latent shape.
The spatial saliency computation requires only a single backward
pass through the reward model per training step, introducing
negligible computational overhead relative to the diffusion model
forward and backward passes, and adds zero cost at inference time.

%% file: sec/5_exps.tex
\section{Experiments}

\begin{figure}[H]
    \centering
    \includegraphics[width=1\linewidth]{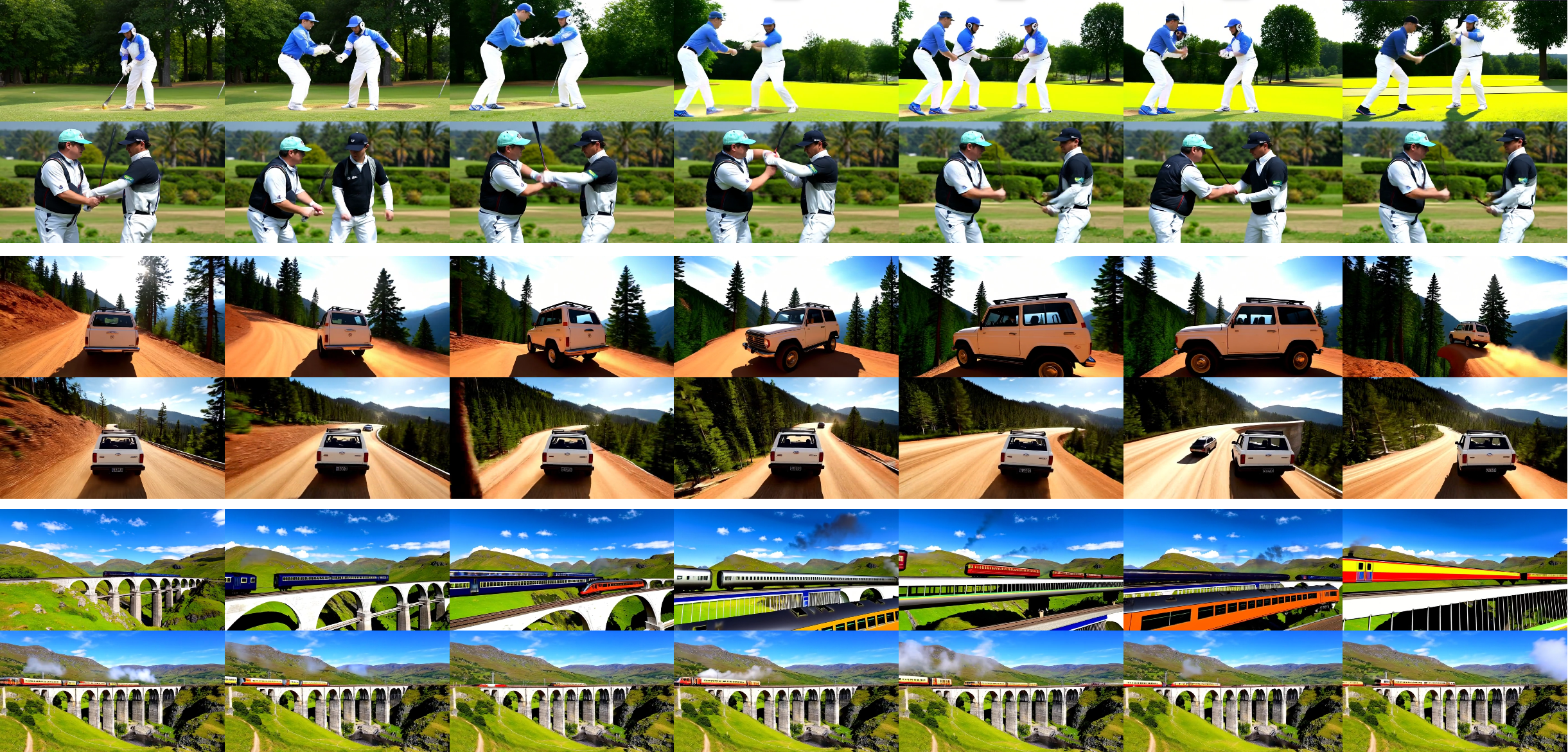}
  \caption{Qualitative comparison on long video generation. For each pair, the top row is Reward Forcing and the bottom row is Stream-R1.}
  \label{fig:visiual2}
\end{figure}

\label{sec:Experiments}
\subsection{Implementation Details}
\label{sec:sub4.1}

Stream-R1 is built upon the Reward Forcing~\cite{lu2025reward} framework, using Wan2.1-T2V-1.3B~\cite{wan2025wan} as the student generator and Wan2.1-T2V-14B as the teacher model to generate 5-second videos at $832 \times 480$ resolution. The model is initialized from a pretrained ODE regression checkpoint trained on 16k ODE solution pairs sampled from the base model, following CausVid~\cite{lin2025autoregressive}. Text prompts for training are drawn from the filtered VidProM dataset, augmented with LLM-based prompt rewriting. Denoising is performed chunk-wise using 3 latent frames per chunk, with denoising steps set to $[1000, 750, 500, 250]$ and an attention window size of 9.

For Stream-R1-specific components, we enable gradient-based spatial saliency computation at the pixel level across all three quality dimensions (VQ, MQ, TA) with adaptive combination ($\tau = 1.0$). Factored spatiotemporal decomposition is applied with spatial minimum weight $\sigma_{\min} = 0.15$ and temporal minimum weight $\tau_{\min} = 0.20$. The reward mode is set to Overall (average of VQ, MQ, TA) with inverse temperature $\beta = 2.0$.

Training runs for 1{,}000 optimizer steps on 8 A100 GPUs with a per-GPU batch size of 1 and gradient accumulation over 8 steps, yielding an effective batch size of 64. The AdamW optimizer is adopted with learning rates of $2.0 \times 10^{-6}$ for the generator $G_\theta$ and $4.0 \times 10^{-7}$ for the fake score $s_{\mathrm{fake}}$, updating the generator every 5 steps and adjusting the fake score accordingly. EMA is applied with a decay weight of 0.99 starting from step 200. The total training time is approximately 56 hours.

\subsection{Comparison with State-of-the-Art}
\label{sec:sub4.2}

\textbf{Short video generation.} We generate 5-second videos
using 946 official VBench~\cite{huang2024vbench} prompts, rewritten
using Qwen2.5-7B-Instruct~\cite{hui2024qwen2} following Self
Forcing~\cite{huang2025self}, each sampled with 5 different seeds
for comprehensive quality assessment. We benchmark our method
against representative open-source video generation models of
comparable scale, including diffusion-based methods (LTX-Video~\cite{hacohen2024ltx},
Wan2.1~\cite{wan2025wan}), autoregressive and streaming models (SkyReels-V2~\cite{chen2025skyreels}, MAGI-1~\cite{teng2025magi},
NOVA~\cite{deng2024autoregressive}, Pyramid Flow~\cite{jin2024pyramidal}, CausVid~\cite{lin2025autoregressive}, Self Forcing~\cite{huang2025self}, LongLive~\cite{yang2025longlive}, Rolling
Forcing~\cite{liu2025rolling}), and reward-guided distillation (Reward Forcing)~\cite{lu2025reward}.
\begin{table}[!htbp]
\centering
\small
\renewcommand{\arraystretch}{1.15}
\setlength{\tabcolsep}{6pt}
\footnotesize
{%
\begin{tabular}{@{}l c r ccc@{}}
\toprule
 & & & \multicolumn{3}{c}{\textbf{VBench Scores} $\uparrow$} \\
\cmidrule(lr){4-6}
\textbf{Model} & \textbf{Params} & \textbf{FPS}$\uparrow$ & Total & Quality & Semantic \\
\midrule
\multicolumn{6}{@{}l}{\textit{Diffusion models}} \\[2pt]
LTX-Video~\cite{hacohen2024ltx}         & 1.9B & 8.98  & 80.00 & 82.30 & 70.79 \\
Wan2.1~\cite{wan2025wan}             & 1.3B & 0.78  & 84.26 & \textbf{85.30} & 80.09 \\[3pt]
\multicolumn{6}{@{}l}{\textit{Autoregressive / streaming models}} \\[2pt]
SkyReels-V2~\cite{chen2025skyreels}        & 1.3B & 0.49  & 82.67 & 84.70 & 74.53 \\
MAGI-1~\cite{teng2025magi}             & 4.5B & 0.19  & 79.18 & 82.04 & 67.74 \\
NOVA~\cite{deng2024autoregressive}               & 0.6B & 0.88  & 80.12 & 80.39 & 79.05 \\
Pyramid Flow~\cite{jin2024pyramidal}       & 2B   & 6.7\phantom{0}   & 81.72 & 84.74 & 69.62 \\
CausVid~\cite{lin2025autoregressive}            & 1.3B & 17.0\phantom{0}  & 82.88 & 83.93 & 78.69 \\
Self Forcing~\cite{huang2025self} & 1.3B & 17.0\phantom{0}  & 83.80 & 84.59 & 80.64 \\
LongLive~\cite{yang2025longlive}           & 1.3B & 20.7\phantom{0}  & 83.22 & 83.68 & \underline{81.37} \\
Rolling Forcing~\cite{liu2025rolling}    & 1.3B & 17.5\phantom{0}  & 81.22 & 84.08 & 69.78 \\[3pt]
\multicolumn{6}{@{}l}{\textit{Reward-guided distillation}} \\[2pt]
Reward Forcing~\cite{lu2025reward}     & 1.3B & 23.1\phantom{0}  & \underline{84.13} & 84.84 & 81.32 \\
\rowcolor{gray!12}
\textbf{Stream-R1 (Ours)} & 1.3B & \textbf{23.1}\phantom{0} & \textbf{84.40} & \underline{85.14} & \textbf{81.44} \\
\bottomrule
\end{tabular}%
}

\caption{\textbf{Short video performance comparison with baselines.} Best results in \textbf{bold}, second-best \underline{underlined}. All methods generate 5-second videos at $832\times480$. Among autoregressive and streaming models, Stream-R1 achieves the highest Quality score. Notably, despite being distilled into a 4-step model, Stream-R1 surpasses its own multi-step diffusion teacher Wan2.1 in both Total and Semantic scores, demonstrating that reward-guided distillation can push the student beyond the teacher's quality frontier.}
\label{tab:main_comparison}

\end{table}
As shown in Tab.~\ref{tab:main_comparison}, Stream-R1 achieves
the highest Total score of 84.40 among all compared methods,
surpassing both the full multi-step diffusion baseline Wan2.1
(84.26) and the previous best distilled model Reward Forcing
(84.13). Among all autoregressive and streaming models,
Stream-R1 attains the highest Quality score (85.14), and on
the Semantic axis, Stream-R1 achieves the best score of 81.44
across all methods, outperforming LongLive (81.37) and Reward
Forcing (81.32).

Compared directly against Reward Forcing, which applies a global
scalar reward and serves as our baseline, Stream-R1 improves
Total, Quality, and Semantic scores simultaneously (+0.27, +0.30,
+0.12), validating that spatiotemporal reward localization enables
more targeted optimization without any additional inference cost.
The Quality improvement is particularly noteworthy: Stream-R1
closes over 60\% of the gap between Reward Forcing (84.84) and
the full diffusion teacher Wan2.1 (85.30).

A striking observation is that Stream-R1, as a 4-step distilled
model, surpasses its own multi-step diffusion teacher Wan2.1 in
Total (84.40 vs.\ 84.26) and Semantic (81.44 vs.\ 80.09) scores
while running at $30\times$ higher inference speed. This challenges
the conventional view that distillation inevitably trades quality
for speed. Through reward-guided distribution matching, the
student is not merely compressed toward the teacher's output
distribution, but actively steered toward higher-reward regions
of the generation manifold. Stream-R1 further amplifies this
effect by concentrating the reward signal on high-perplexity regions, enabling the optimizer to discover quality
improvements that uniform global weighting overlooks.

\textbf{Long video generation.}
Following the evaluation protocol of Reward Forcing~\cite{lu2025reward},
we generate videos at five durations (10s, 30s, 60s, 120s, 180s) using
the first 128 prompts from MovieGen Video Bench with autoregressive
block-wise generation and EMA-Sink attention.

\begin{figure*}[!htbp]
    \centering
    \includegraphics[width=\textwidth]{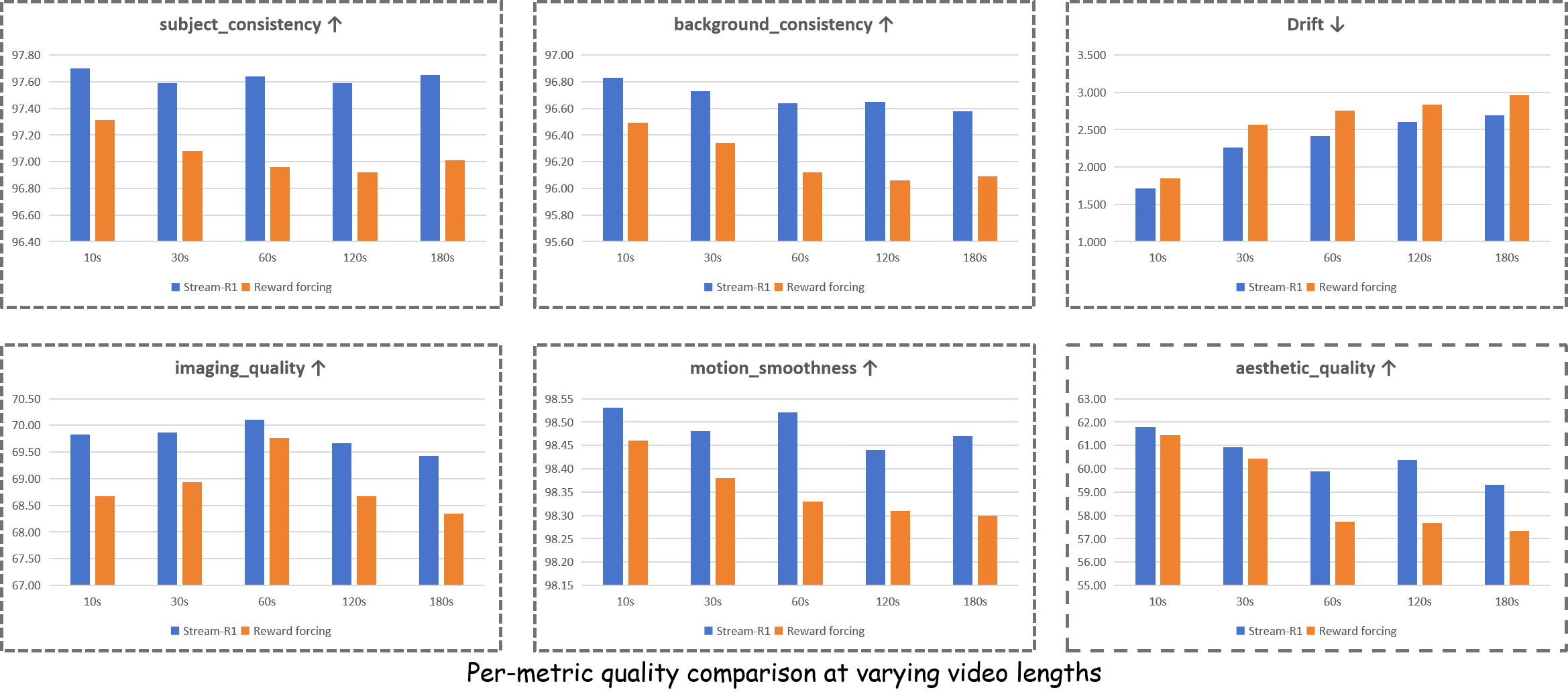}
    \vspace{-2mm}
    \caption{\textbf{Per-metric quality comparison at varying video lengths.}
    Stream-R1 (blue) consistently outperforms Reward Forcing (orange) across
    all six metrics at every duration. The advantage widens as video length
    increases, particularly at 120s and 180s, confirming that spatiotemporal
    reward-guided weighting mitigates the quality drift accumulated during
    long autoregressive rollouts.}
    \label{fig:long_video_scaling}
\end{figure*}

As shown in Fig.~\ref{fig:long_video_scaling}, Stream-R1 outperforms
Reward Forcing on all six VBench metrics across every evaluated duration.
Two observations stand out. First, Stream-R1 achieves consistently higher
absolute scores on subject consistency, background consistency, imaging
quality, motion smoothness, and aesthetic quality, while maintaining
lower drift throughout. Second, and more importantly, the performance gap
\textit{widens} as the video length grows: at 10s the two methods are
relatively close, but by 120s and 180s Stream-R1 retains notably higher
quality while Reward Forcing degrades more steeply. This widening gap
confirms that spatiotemporal reward-guided weighting not only improves
per-frame quality but also slows the rate of quality collapse along the
temporal axis, a direct benefit of the temporal weighting component that
prevents artifacts from propagating into subsequent chunks.

\textbf{VLM-based evaluation.}
Following Reward Forcing~\cite{lu2025reward}, we additionally employ
Qwen3-VL-235B-A22B-Instruct~\cite{bai2025qwen3} to evaluate long
video generation quality, scoring each of the 128 videos from 1 to 5
on visual quality, motion dynamics, and text alignment. As shown in
Tab.~\ref{tab:vlm_long}, Stream-R1 attains the highest Visual Quality
and Text Alignment among all compared methods, with overall scores
that remain competitive across the three axes. This balanced profile
is consistent with the design of the multi-dimensional reward, which
distributes optimization across visual quality, motion, and text
alignment rather than concentrating on any single one.
\begin{table}[!htbp]
\centering
\caption{\textbf{VLM-based evaluation on 60-second videos.}
Qwen3-VL scores across three quality dimensions.
Best in \textbf{bold}, second-best \underline{underlined}.}
\label{tab:vlm_long}
\vspace{-2mm}
\small
\begin{tabular}{l ccc}
\toprule
Model & Visual\,$\uparrow$ & Dynamic\,$\uparrow$ & Text\,$\uparrow$ \\
\midrule
SkyReels-V2~\cite{chen2025skyreels}
  & 3.30 & 3.05 & 2.70 \\
CausVid~\cite{lin2025autoregressive}
  & \underline{4.66} & 3.16 & 3.32 \\
Self Forcing~\cite{huang2025self}
  & 3.89 & 3.44 & 3.11 \\
LongLive~\cite{yang2025longlive}
  & 4.79 & 3.81 & 3.98 \\
Reward Forcing~\cite{lu2025reward}
  & 4.82 & \textbf{4.18} & \underline{4.04} \\
\rowcolor{gray!12}
Stream-R1 (Ours)
  & \textbf{4.92} & \underline{4.04} & \textbf{4.11} \\
\bottomrule
\end{tabular}
\end{table}

\textbf{Human preference evaluation.}
To complement automated metrics, we conduct a human preference study
on 50 long videos (60\,s). Annotators are presented with
anonymized A/B pairs from Reward Forcing and Stream-R1 and asked to
judge five dimensions: temporal consistency, dynamic reasonableness,
visual quality, text alignment, and overall preference.
As shown in Tab.~\ref{tab:human_eval}, Stream-R1 is preferred on
all five dimensions, with the largest margins on Dynamic
Reasonableness (63.0\%) and Visual Quality (60.0\%). Human judgment
provides a complementary perspective to automated metrics, since
flow-based dynamics scores do not distinguish camera motion from
subject motion, while perceived motion quality is what evaluators
assess directly.

\begin{table}[H]
\centering
\small
\caption{\textbf{Human preference evaluation.} Win rate (\%) of
Stream-R1 vs.\ Reward Forcing on 50 long videos (60\,s),
judged by 5 annotators. Win Rate $=$ (Win $+$ 0.5 $\times$ Tie) / Total.}
\label{tab:human_eval}
\vspace{-2mm}
\renewcommand{\arraystretch}{1.15}
\begin{tabular}{l ccc c}
\toprule
Dimension & Win & Tie & Lose & Win Rate \\
\midrule
Temporal Consistency  & 25 & 1 & 24 & 51.0\% \\
Dynamic Reasonableness & 30 & 3 & 17 & \textbf{63.0\%} \\
Visual Quality \& Aesthetics & 29 & 2 & 19 & \textbf{60.0\%} \\
Text-Video Alignment  & 22 & 9 & 18 & 54.1\% \\
\midrule
\rowcolor{gray!12}
Overall Preference    & 28 & 1 & 21 & \textbf{57.0\%} \\
\bottomrule
\end{tabular}
\end{table}

\subsection{Ablation study.}
We progressively add each proposed component on top of the
distribution matching distillation baseline (i.e., without reward feedback) and evaluate on both short (VBench 946)
and long (60-second) video benchmarks.
Results are summarized in Tab.~\ref{tab:ablation}.
\definecolor{accent}{HTML}{2E86AB}

\begin{table}[!htbp]
\centering
\small
\caption{\textbf{Ablation study of Stream-R1 components.}
Each row adds one component to the baseline.
$\sigma_{\min}$: spatial weight floor; $\tau_{\min}$: temporal weight floor.}
\label{tab:ablation}
\vspace{-2mm}
\renewcommand{\arraystretch}{1.15}
{%
\begin{tabular}{l ccc cc}
\toprule
\multirow{2}{*}{Variant}
  & \multicolumn{3}{c}{Short Video (VBench)\,$\uparrow$}
  & \multicolumn{2}{c}{Long Video\,$\uparrow$} \\
\cmidrule(lr){2-4} \cmidrule(lr){5-6}
  & Total & Quality & Semantic & Total & Drift\,$\downarrow$ \\
\midrule
Baseline
  & 83.44 & 84.16 & 80.55 & 79.45 & 2.479 \\
+ Spatial reward ($\sigma_{\min}\!=\!0.15$)
  & 83.67 & 84.46 & 80.51 & 80.71 & 2.653 \\
+ Balanced Multi-Dim reward ($\sigma_{\min}\!=\!0.15$)
  & 83.67 & 84.45 & 80.54 & 80.72 & 2.651 \\
\rowcolor{gray!12}
+ Temporal reward ($\tau_{\min}\!=\!0.20$) \textit{[Full]}
  & \textbf{84.40} & \textbf{85.14} & \textbf{81.44} & \textbf{80.86} & \textbf{2.417} \\
\midrule
\multicolumn{6}{l}{\textit{Hyperparameter sensitivity}} \\
\quad $\sigma_{\min}=0.30$ (spatial only)
  & 83.68 & 84.44 & 80.62 & 80.73 & 2.697 \\
\quad $\tau_{\min}=0.40$
  & 83.42 & 84.21 & 80.24 & 80.40 & 2.475 \\

\bottomrule
\end{tabular}
}
\end{table}

\textit{Effect of spatial saliency.}
Adding gradient-based spatial saliency weighting to the baseline
improves Quality from 84.16 to 84.46 and Long Total from 79.45 to
80.71.

\textit{Effect of balanced multi-dimensional reward.}
Replacing single-metric saliency with the adaptive balanced
combination yields a Semantic improvement (80.51$\to$80.62)
while maintaining comparable Quality, indicating that the
balanced scheme prevents any single reward dimension from
dominating the spatial weighting.

\textit{Effect of temporal decomposition.}
Incorporating temporal saliency decomposition produces the
largest single improvement: Short Total jumps from 83.68 to
84.40 (+0.72), and Drift drops from 2.697 to 2.417.

\textit{Sensitivity to temporal floor $\tau_{\min}$.}
Setting $\tau_{\min}=0.40$ (versus the default $0.20$) degrades
Short Total from 84.40 to 83.42, even below the spatial-only
variant. An excessively high temporal floor suppresses the contrast between 
high-saliency and low-saliency frames, effectively reducing 
temporal saliency to uniform weighting and forfeiting its benefit.

\subsection{Visualization of spatiotemporal weights.}

To probe whether the spatiotemporal weights actually respond to local quality deficiency, we conduct a controlled visualization that introduces contrast at \emph{two} levels. Within each frame, we inject a localized Gaussian blur only into the lower half, leaving the upper half intact, so that every single frame contains a clean region and a degraded region side by side. Across frames, we hold the blur intensity fixed but progressively enlarge the corrupted area from left to right. We then back-propagate the reward score through the vision encoder to obtain the real gradient saliency.

\begin{figure}[H]
    \centering
    \includegraphics[width=1\linewidth]{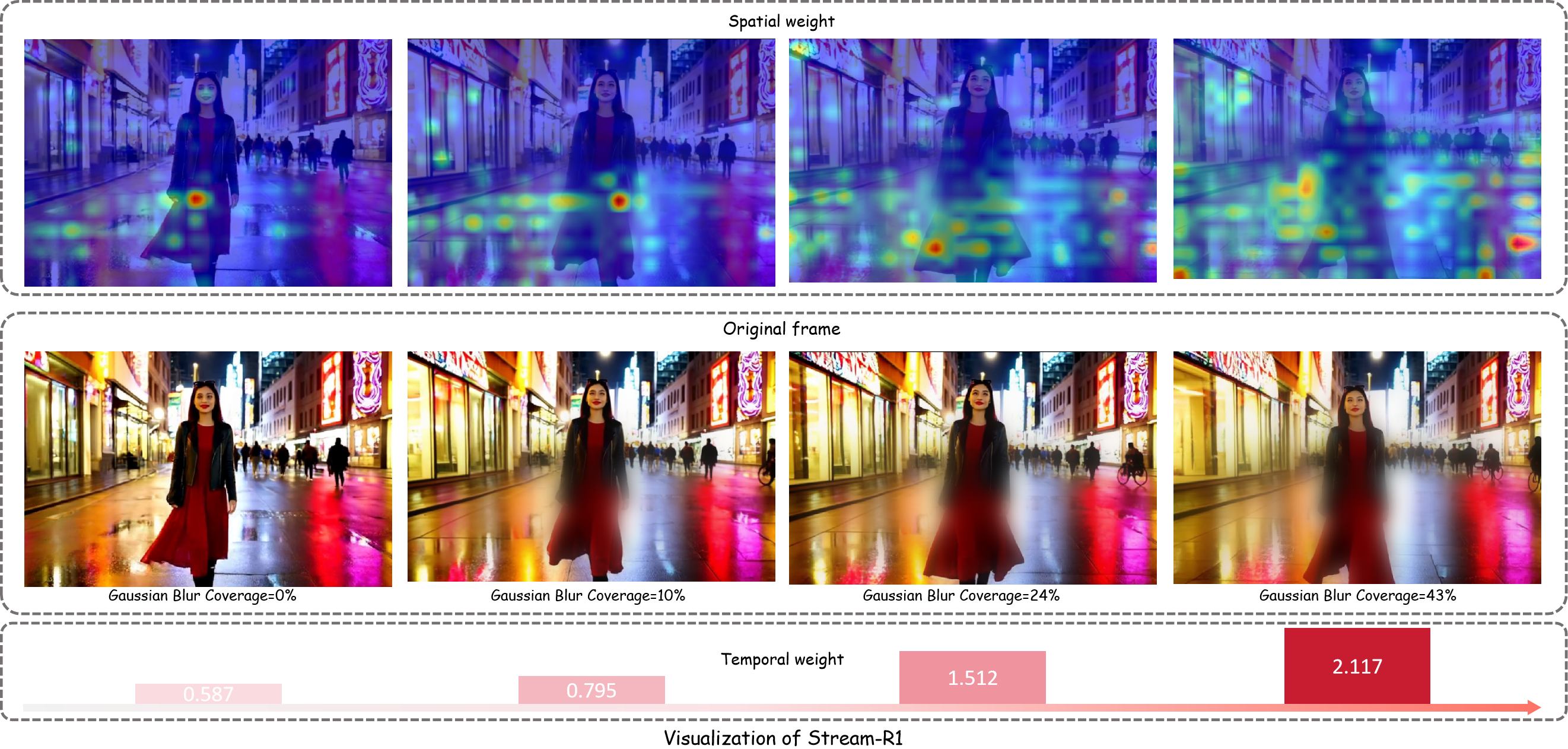}
  \caption{Spatiotemporal saliency under controlled degradation. Gaussian blur is injected only into the lower half of each sampled frame so that every frame itself forms a clean (top) versus degraded (bottom) contrast; the blurred area further expands across the four frames from left to right. \textbf{Top:} reward-model gradient saliency overlaid on the degraded frames. \textbf{Middle:} the degraded frames, where only the lower half is corrupted. \textbf{Bottom:} per-frame temporal weights $w_t$, growing as the degraded area enlarges.}
  \label{fig:visiual}
\end{figure}

Fig.~\ref{fig:visiual} reports the result. In the leftmost frame, where the injected blur covers only a small lower patch, the saliency naturally concentrates on the human face and on the lower-body region with strong motion cues, suggesting that even in a relatively clean frame the reward gradient tends to emphasize semantically and dynamically important content. As the blurred region grows across the subsequent frames, the saliency progressively migrates toward the enlarged corrupted area and tightens onto its interior. Within each individual frame, activations are also clearly biased toward the lower (blurred) half rather than the visually intact upper half, indicating that the reward-model gradient tends to highlight regions where quality refinement would yield the greatest perceptual gain rather than regions that are already clean.

The bottom row shows the corresponding temporal weights, which grow monotonically from 0.587 to 2.117 as the degraded area expands. This indicates that frames containing more quality-deficient content are automatically up-weighted by the temporal aggregation, allocating more learning signal to the frames that need it most. Importantly, this behavior is not hand-engineered: it emerges purely from gradients of the reward model, providing direct evidence that Stream-R1's spatiotemporal weights reflect localized quality deficiency in both space and time.

Fig.~\ref{fig:visiual2} further provides qualitative comparisons on long video generation, where the top row of each pair is Reward Forcing and the bottom row is Stream-R1. Stream-R1 produces more temporally consistent appearance, stable backgrounds, and coherent motion, while Reward Forcing shows visible drift and deformation over time.

%% file: sec/7_conclusion.tex
\section{Conclusion}
\label{sec:Conclusion}
We present Stream-R1, a dynamic spatiotemporal reward-guided 
distillation framework that decomposes scalar reward signals into 
factored spatial and temporal saliency via reward-model gradient 
backpropagation, concentrating optimization intensity on 
quality-deficient regions and frames at no additional inference 
cost. Experiments show that Stream-R1 achieves the highest overall 
VBench score among all compared methods, including its multi-step 
bidirectional teacher Wan2.1. On long video generation, Stream-R1 
attains the best imaging quality and lowest drift, demonstrating 
superior temporal stability. Both VLM-based and human preference 
evaluations confirm that Stream-R1 delivers balanced quality 
improvements: it achieves the highest VLM visual quality and text 
alignment scores, and is preferred by human evaluators on all five 
judged dimensions, with particularly strong advantages on dynamic 
reasonableness and visual quality. We believe the principle of 
decomposing global reward signals into spatiotemporally localized 
guidance, rather than treating reward as a monolithic scalar, opens 
a new direction for reward-guided generation and is broadly 
applicable to other modalities including image synthesis and 3D 
content generation.